\documentclass{article}
\usepackage[preprint]{spconf}
\usepackage{amsmath,graphicx,lipsum,xr}
\ninept

\newlength{\figskip}
\title{Improved TDNNs using Deep Kernels and Frequency Dependent Grid-RNNs}
\name{F. L. Kreyssig\footnote{Supported by Studienstiftung des Deutschen Volkes}, C. Zhang, P. C. Woodland\thanks{Thanks to Mark Gales and the MGB3 team for the MGB3 setup used. Florian Kreyssig is supported by the Studienstiftung des Deutschen Volkes.}}
\address{Cambridge University Engineering Dept., Trumpington St., Cambridge, CB2 1PZ U.K.\\
{\small \tt \{flk24,cz277,pcw\}@eng.cam.ac.uk}
}

\begin{document}
%
\maketitle
\begin{abstract}
Time delay neural networks (TDNNs) are an effective acoustic model for large vocabulary speech recognition. The strength of the model can be attributed to its ability to effectively model long temporal contexts. However, current TDNN models are relatively shallow, which limits the modelling capability. This paper proposes a method of increasing the network depth by deepening the kernel used in the TDNN temporal convolutions. The best performing kernel consists of three fully connected layers with a residual (ResNet) connection from the output of the first to the output of the third. The addition of spectro-temporal processing as the input to the TDNN in the form of a convolutional neural network (CNN) and a newly designed Grid-RNN was investigated. The Grid-RNN strongly outperforms a CNN if different sets of parameters for different frequency bands are used and can be further enhanced by using a bi-directional Grid-RNN. Experiments using the multi-genre broadcast (MGB3) English data (275h) show that deep kernel TDNNs reduces the word error rate (WER) by 6\% relative and when combined with the frequency dependent Grid-RNN gives a relative WER reduction of 9\%.
\end{abstract}
\begin{keywords}
Time Delay Neural Network, Grid Recurrent Neural Network, Speech Recognition, ResNet
\end{keywords}
\toappear{To appear in Proc. ICASSP 2018, April 15-20, 2018, Calgary, Canada}
\copyrightnotice{\copyright IEEE 2018}
\section{Introduction}
\label{sec:intro}
Artificial Neural Networks have become the dominant approach to acoustic modelling, achieving dramatic improvements on a 
range of tasks. One commonly used neural network architecture is the time-delay neural network (TDNN), originally proposed in \cite{waibel1989phoneme} and often used in its \textit{sub-sampled} form \cite{peddinti2015time}. A TDNN consists of identical fully-connected (FC) layers repeated at different time-steps. It is thus seen as a forerunner to the convolutional neural network (CNN) \cite{lecun1998gradient}, which applies the FC layer with specified regular shifts. While the TDNN does not have this restriction, the CNN formulation extends easily to multiple axes e.g. time and frequency. Shifting the FC layer incorporates the assumption that the same important features can occur at various time-steps. Incorporating this knowledge reduces the computation and number of parameters required. It is also known that a difference in speaker can express itself in small shifts in frequency. This knowledge can be incorporated into the model by applying CNNs along the frequency dimension \cite{abdel2012applying}, 
which will be one extension to the TDNN that this work evaluates.

Deepening neural networks continues to yield improvements in performance. A striking example in computer vision  is the improvement on the ImageNet classification task when transitioning from AlexNet \cite{krizhevsky2012imagenet} (8 Layers), to VGG \cite{Simonyan15} (19 Layers) to ResNet \cite{he2016deep} (152 Layers). These were all based on CNNs and the increase in depth came from stacking more convolutional layers. 
This work proposes a method of deepening a TDNN, similar to the one proposed in \cite{lin2013network}, which is to make the kernel used for the temporal convolution deep. This results in a potentially much more complex kernel.

Two dimensional recurrent neural networks (RNNs) are gaining popularity over CNNs for the modelling of spectro-temporal variations \cite{exploring-multidimensional-lstms-for-large-vocabulary-asr,Tara2016ModelingTF,Li2017AcousticGoogleHome,li2017reducing}.  Here,  an efficient Grid-RNN is designed, which can be used as the input to the TDNN architecture and uses a separate set of parameters for different frequency bands.
Experiments on the multi-genre broadcast English (MGB3) challenge  task are used to evaluate the different changes in architecture and demonstrate the efficacy of the proposed improvements.

The remainder of this paper is organised as follows. Section 2 outlines the proposed extensions to the sub-sampled TDNN architecture and discusses related work. Sections 3 and 4 present the experimental setup and results and Sec. 5 gives conclusions.

\section{Models Investigated}
\label{sec:Models}

\subsection{Time-Delay Neural Networks}
\label{ssec: TDNN}
A TDNN starts with an FC layer that takes a stack of frames as its input and is replicated across different time-steps. The following layer then also takes as input a stack of different time-steps of the preceding layer and is also replicated across different time-steps. The initial layers thus learn to detect features within narrow temporal contexts while the later layers operate on a much larger temporal context.

\begin{figure}[htb]
  \centering
  \centerline{\includegraphics[width=8.3cm]{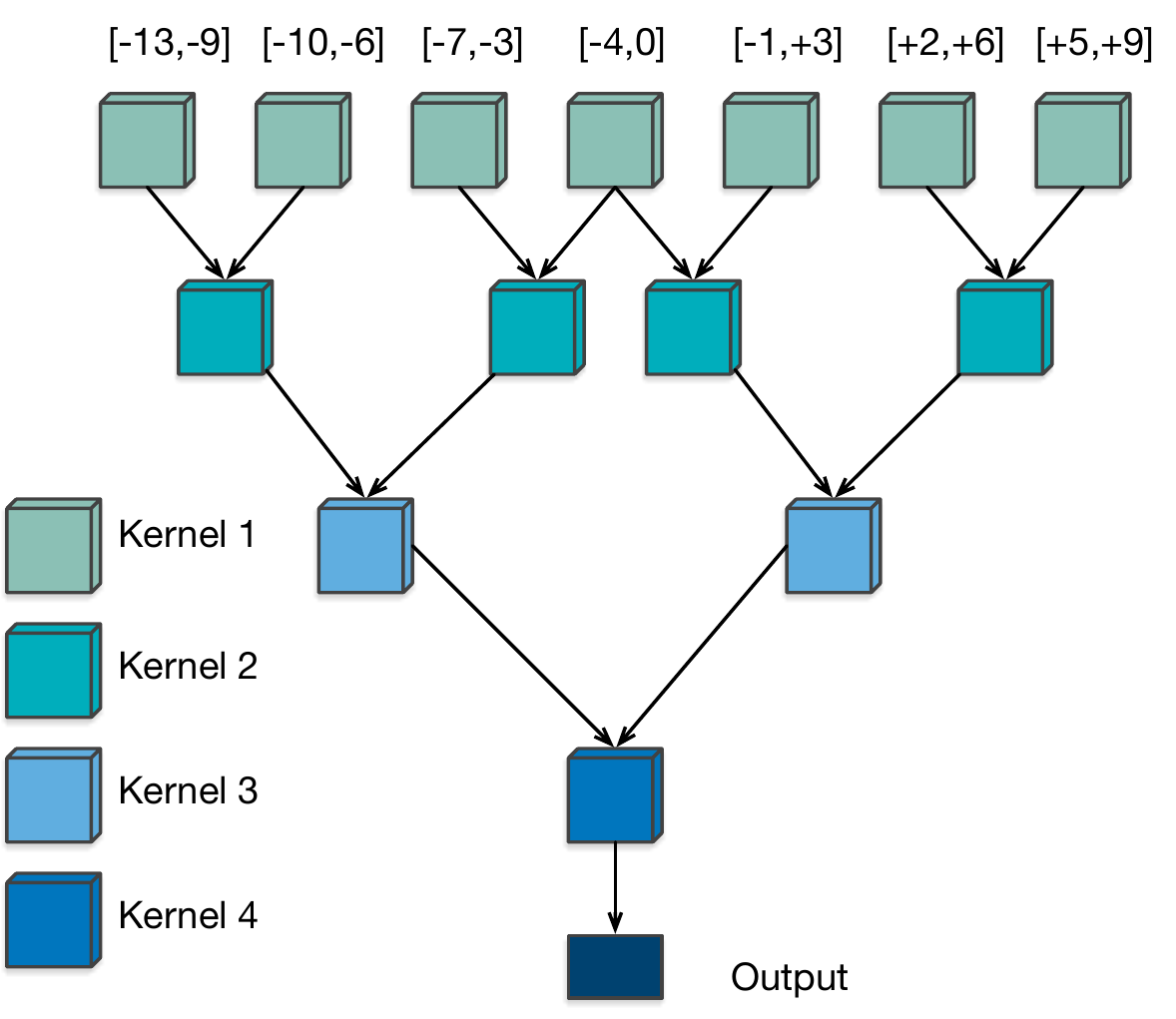}}
  \vspace{\figskip}
  \caption{Baseline TDNN Structure}
\label{fig:BaseTDNN}
\end{figure}

The original TDNN formulation \cite{waibel1989phoneme,waibel1989modular} uses shifts of one frame in time for the FC layers, which is very expensive both in terms of computation and the numbers of parameters when operating on large temporal contexts. It was shown in \cite{peddinti2015time}  that neither one frame shifts  nor uniform time shifts are necessary. Their proposed sub-sampled TDNN, illustrated in Fig. \ref{fig:BaseTDNN}, is constructed by moving the first FC layer across the window $t \in [-13,9]$ with a temporal context of 5 frames at shifts of 3 frames, thus splitting the total input context into 7 time-bins. This is followed by a binary tree combination of the outputs of the layer. This TDNN will be the baseline acoustic model in this paper. 
During back-propagation,  gradients are accumulated over all instantiations of the FC layers and then normalised.

\subsection{Deep Kernels}
\label{ssec: DeepKernels}
The sub-sampled TDNN has an effective and well-studied structure. However, the structure of sub-sampled TDNNs generally limits their depth, which in turn limits their modelling strength. Hence, in order to keep to the underlying structure, we try to increase the modelling capability of each of the kernels. For image classification, \cite{lin2013network} proposed a CNN structure in which the convolutional filter is replaced by a stack of FC layers. This means that each filter can learn to represent more abstract features.

\begin{figure}[htb]
\centering
\centerline{\includegraphics[width=8.5cm]{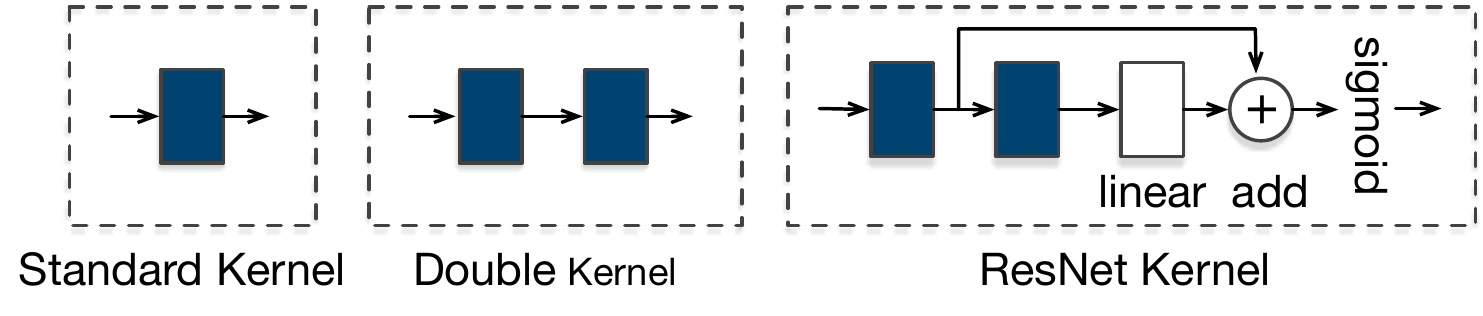}}
\vspace{\figskip}
\caption{Different kernels investigated. Darker blocks are FC layers with $\sigma(\cdot)$ activation function. The white block denotes an FC layer with linear activation function.}
\label{fig:Kernels}
\end{figure}

Three types of temporal kernels, shown in Fig.~\ref{fig:Kernels}, are evaluated in our experiments. The first kernel is the \textit{Standard}-Kernel, a simple FC layer as used in the baseline structure. The second kernel, the \textit{Double}-Kernel, is built by using two FC layers instead of one. Finally, the \textit{ResNet}-Kernel, is constructed  by appending an FC layer with two further FC layers that can be bypassed by a residual connection. Using the ResNet-Kernel increases the depth of the TDNN (including the output layer) from 5 layers to 13.

Even though deeper networks generally improve the performance of neural network architectures, they are usually harder and slower to train. Residual connections \cite{he2016deep} are identity mappings from the output of initial layers to the input of later layers and can be described by: \begin{align}
\mathbf{y = \mathcal{F}\left(\mathbf{x}, \mathbf{\theta}\right) + \mathbf{x}}
\end{align} where $\mathbf{x}$ and $ \mathcal{F}\left(\mathbf{x}, \mathbf{\theta}\right)$ are the input and the output of the block of layers that is to be ``skipped". This direct connection means that back-propagation of the gradients in very deep structures is more effective, since the effective minimum depth, in terms of layers, is reduced.  For the ResNet-Kernel, the effective  minimum depth is reduced from three layers to a single layer. Furthermore, it has been hypothesised that it is simpler to optimise the residual function $\mathcal{F}$ than the combined mapping. The residual function $\mathcal{F}$ in the \textit{ResNet}-Kernel is an FC layer with Sigmoid activation function, $\sigma(\cdot)$, followed by an FC layer with linear activation function. A $\sigma$ activation function cannot be used in the last layer of the  \textit{ResNet}-Kernel since, due to its non-negative output range, it could only increase the input signal. Hence, here a linear activation function is used.

\subsection{Frequency Domain Convolution}
\label{ssec: CNN}
CNNs have shown promising results on a range of speech  tasks \cite{abdel2012applying,sainath2013deep,sainath2015convolutional,sercu2016very,Qian2016VeryDeep,convolutional-neural-networks-for-speech-recognition-2}. In a CNN layer a set of filters is convolved with the input which results in multiple output-maps, one per filter. This is followed by the application of an element-wise activation function, such as the $\sigma(\cdot)$ function. The operation that the layer performs on an input map with two axes, such as a spectrogram (time $\!\times\!$ frequency), can be written as:
\begin{align}
h_{i,j,k} = \sigma\left(\sum^{L-T}_{l = 0} \sum^{M-F}_{m = 0} x_{i+l,j+m} \cdot w_{l,m,k}\right)\!,k = 1 \dots K \label{eq: convolution}
\end{align}
where $L$ is the dimensionality of the time-axis, $M$ is the dimensionality of the frequency-axis, $T\!\times\!F$ is the size of the filters, $k$ is the index of the filter and K is the number of filters. $w_{l,m,k}$ are the learned parameters of the CNN. This is generally followed by a \textit{pooling} operation which ``summarises" patches in each output map by either computing their average (average-pooling) or their maximum value (max-pooling). This allows for some invariance to shifts in the location of a feature. 

In the above description, the same filter would be applied across the entire input space, known as \textit{full weight sharing} (FWS). This assumes that a feature can occur across the entire input space. This is a valid assumption for the temporal axis and hence done in the TDNN architecture. However, it is not generally the case for the frequency axis as the characteristics are different for higher and lower frequencies. To incorporate this into the model design, \textit{limited weight sharing} (LWS) is introduced in \cite{abdel2012applying,convolutional-neural-networks-for-speech-recognition-2}, where a specific filter is used only for a specific part of the frequency axis, the outputs of which are then max-pooled to a single scalar. Initial experiments did not find this method to give significant improvements, possibly due to the strong loss of information from the large pooling sizes. Hence, we introduce a convolution strategy that strikes a balance between FWS and LWS. The frequency axis is divided into different, but overlapping, frequency bands and convolution followed by max-pooling is performed within these frequency bands and the outputs concatenated. In comparison to LWS the pooling operation does not span the entire frequency band. The convolution can then be described by:
\begin{align}
h^{f}_{i,j,k}= \sigma\left(\sum^{L-T}_{l = 0}\;\; \sum^{M+f\cdot S-F}_{m = f\cdot S}\!\!\!\!\!x_{i+l,j+m} \cdot w_{l,m,k}\right)\!,k = 1 \dots K\label{eq: FBANDconvolution}
\end{align}
where $f$ is the index of the frequency band and $S$ is the shift between the frequency bands. The frequency bands are size $M=10$ and overlap by 5, giving $S = 5$. The input is 40-dimensional along the frequency axis resulting in 7 frequency bands. The filter has size $T\times F = 5\times 5$ resulting in a size of $6\times 1$ for the output which is reduced to $3\times 1$ by the max-pooling layer that has pooling size 2.

\subsection{Frequency Dependent Grid-RNN}

\begin{figure}[htb]
  \centering
  \centerline{\includegraphics[width=8.5cm]{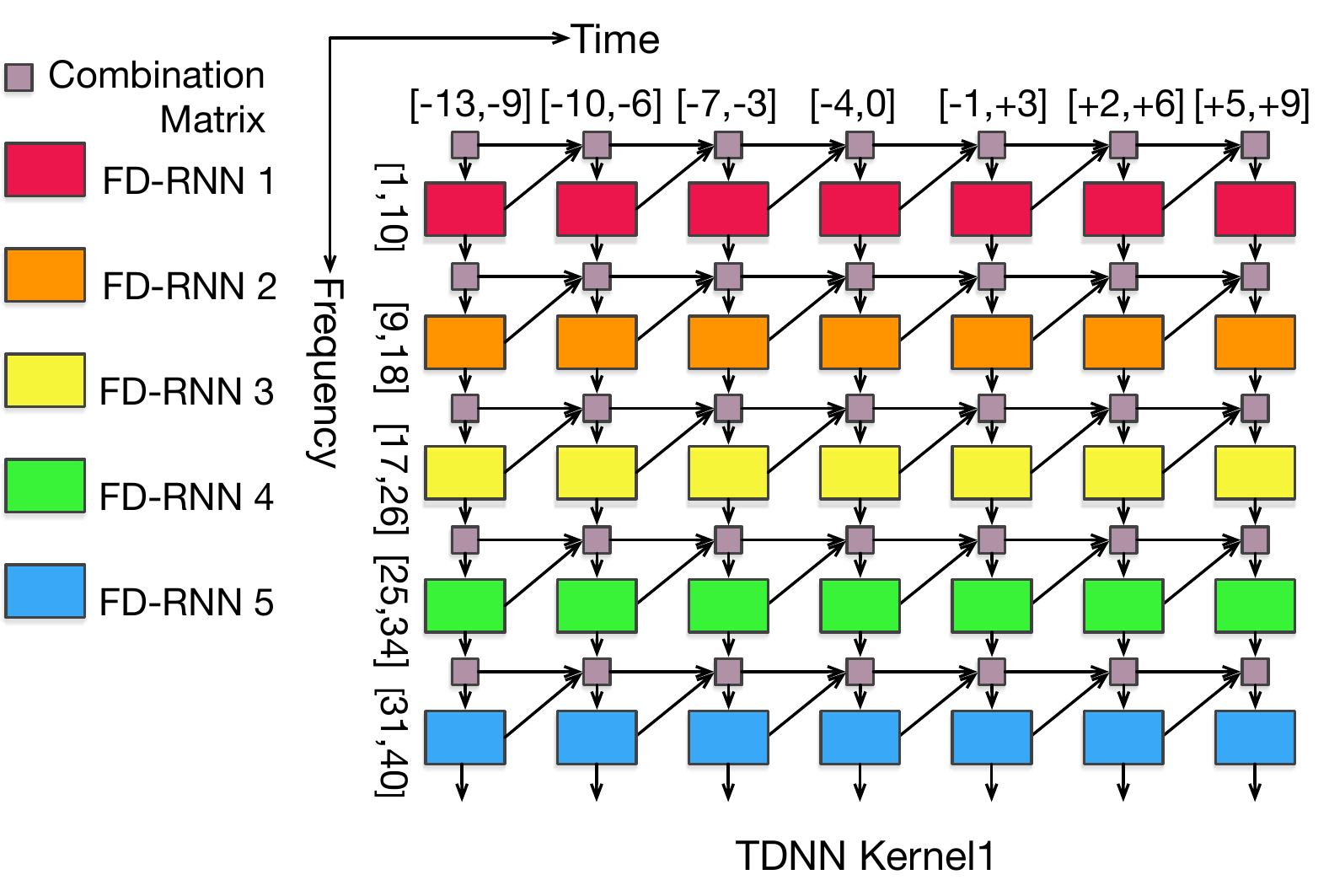}}
\vspace{\figskip}
  \caption{Frequency Dependent Grid-RNN TDNN}
\label{fig:GridRNNTDNN}
\vspace{-9pt}
\end{figure}

RNNs are widely used in speech recognition, often in the form of the Long Short-Term Memory (LSTM) architecture \cite{hochreiter1997long,Sak2014PLSTM,peddinti2017low}. Recently, standard time-domain LSTMs have been extended to model both the time and the frequency dimensions \cite{li2015lstm,exploring-multidimensional-lstms-for-large-vocabulary-asr}. This is done by unfolding the two-dimensional (2D) RNNs along both time and frequency.
This gives the advantage over CNNs of being able to model correlations between features in time and frequency. Grid-LSTMs \cite{kalchbrenner2015grid} have been shown to outperform CNNs as well as the TF-LSTMs \cite{exploring-multidimensional-lstms-for-large-vocabulary-asr} as the input to 1-D LSTM layers \cite{Tara2016ModelingTF}. Both Grid-LSTMs and TF-LSTMs were unfolded in time one time-step at a time and used the computationally expensive LSTM. 

This paper proposes an efficient Grid-RNN architecture, shown in Fig.~\ref{fig:GridRNNTDNN} that uses the vanilla-RNN and groups the input window ($t \in \left[-13,9\right]$) into the same seven time-bins as described in Sec.~\ref{ssec: TDNN}. Thus, it can be neatly combined with the previously discussed TDNN architectures.
The frequency axis is split into five different bins of size ten, which are shown in Fig.~\ref{fig:GridRNNTDNN}. The Grid-RNN structure is 
composed of two RNNs, one with the $\sigma(\cdot)$ activation function and one with a linear activation. The $\sigma$-RNN ($\mathbf{h}_{t,k}^F$) performs feature extraction and the \textit{Linear}-RNN ($\mathbf{h}_{t,k}^I$), called the \textit{Combination Matrix} in Fig.~\ref{fig:GridRNNTDNN}, models the information flow between instantiations of the $\sigma$-RNN. The linear activation function is used for its improved flow of information over the $\sigma(\cdot)$ activation function. The structure is trained in the unfolded form shown, similar to the work done in \cite{Saon2014Unfolded}. The Grid-RNN is described by the following equations:
\begin{align}
\mathbf{h}^{I}_{t,k} &= V^{I}_{F} \mathbf{h}^{F}_{t,k-1} + V^{I}_{I} \mathbf{h}^{I}_{t-1,k} + \mathbf{b}^{I} \label{eq: 2DRNN1}\\
\mathbf{h}^{F}_{t,k} &= \sigma \left( W^{F}_{\left(k\right)} \mathbf{x}_{t,k} + V^{F}_{\left(k\right)} \mathbf{h}^{I}_{t,k-1} + \mathbf{b}^{F}_{(k)}  \right) \label{eq: 2DRNN2}
\end{align}
where $W^{F}_{\left(k\right)}$ is the frequency-dependent (FD) input weight-matrix, $V^{F}_{\left(k\right)}$ is the FD recurrent weight-matrix and $V^{I}_{\alpha}$ are the weight-matrices modulating the information flow. Here, $t$ is the time-band and $k$ is the frequency-band. 

For comparison we can formulate vanilla-RNN versions of the TF-LSTM, which we coin TF-RNN,
\begin{align}
\mathbf{h}_{t,k} &= \sigma\left(W \mathbf{x}_{t,k} + V_{T} \mathbf{h}_{t-1,k} + V_{F} \mathbf{h}_{t,k-1} + \mathbf{b}\right) \label{eq: TFRNN}
\end{align}
and the 2D-Grid-LSTM, which uses two separate LSTMs to model the correlations in frequency and in time:
\begin{align}
\mathbf{h}^{T}_{t,k} &= \sigma\left(W^{T} \mathbf{x}_{t,k} + V_{T} \mathbf{h}^{T}_{t-1,k} + V_{F} \mathbf{h}^{F}_{t,k-1} + \mathbf{b}^{T}\right) \label{eq: GridRNN1}\\
\mathbf{h}^{F}_{t,k} &= \sigma\left(W^{F} \mathbf{x}_{t,k} + V_{T} \mathbf{h}^{T}_{t-1,k} + V_{F} \mathbf{h}^{F}_{t,k-1} + \mathbf{b}^{F}\right) \label{eq: GridRNN2}
\end{align}

From Eqn.~\ref{eq: TFRNN} it can be seen that the TF-RNN is equivalent to the proposed structure if the \textit{Linear}-RNN of Eqn.~\ref{eq: 2DRNN1} is replaced by a concatenation of the two inputs $\mathbf{h}^{F}_{t,k-1}$ and $\mathbf{h}^{I}_{t-1,k}$. This structure would have a much longer series of non-linear mappings leading to a potentially very strong loss of information as it moves through the network. By comparison the \textit{Linear}-RNN provides a linear path in time through the network. This is partly analogous to the ``skip" connections in time within the TF-LSTM or in frequency and time for the Grid-LSTM that is provided by the memory cell.

Using the same reasoning as in Sec.~\ref{ssec: CNN}, it should be beneficial to untie the parameters of the $\sigma$-RNN along the frequency axis, denoted by a frequency dependent Grid-RNN (FD-Grid-RNN). This is shown by the colour-coding in Fig.~\ref{fig:GridRNNTDNN} and by the potentially frequency-band-specific weights in Eqn.~\ref{eq: 2DRNN2}. 

The Grid-RNN has an inherent directionality from the past to the future and from lower frequencies to higher frequencies. The recurrent weight-matrices $V^{\alpha}_{\beta}$ (Eqn.~\ref{eq: 2DRNN1}-\ref{eq: 2DRNN2}) of the two RNNs are likely to have a spectral radius below 1, especially due to our use of L2-regularisation. This means that any information provided to the model vanishes exponentially fast \cite{pascanu2013difficulty}. Thus, it is hypothesised that the model can be improved by using a bi-directional FD-Grid-RNN (BD-FD-Grid-RNN), which is constructed by training two FD-Grid-RNNs in parallel, one with the directionality as in 
Fig.~\ref{fig:GridRNNTDNN} and one with the directions along both the time and frequency axes reversed. The outputs of both are concatenated. This bi-directionality does not increase the inherent latency of the model due its unfolded structure.

\section{Experimental Setup}
\label{sec: ExpSetup}
The proposed architectures were evaluated using multi-genre broadcast (MGB) data \cite{Bell2015MGB} from the MGB3 speech recognition challenge task \cite{MGB2017WEB}. A 275 hour (275h) training set was selected from 750 episodes where the sub-titles have a phone matched error rate $<~40\%$ compared with the lightly supervised output \cite{lanchatin2015g} which was used as training supervision. A 55 hour (55h) subset was sampled at the utterance level from the 275h set. A 63k word vocabulary \cite{Richmond2010Combilex} was used and a trigram word level language model (LM) estimated from both the acoustic transcripts and a separate 640 million word MGB subtitle archive. The test set, $\textbf{dev17b}$, contains 5.55 hours of audio data and 5,201 manually segmented utterances from 14 episodes of 13 shows. 
System outputs were evaluated  with confusion network decoding (CN)  \cite{Mangu2000CN,Evermann2000CN} as well as 1-best Viterbi decoding.

All experiments were conducted with an extended version of HTK 3.5 \cite{Young2015HTK,zhang2015general}. A 40d log-Mel filterbank (FBK) analysis was used without any delta coefficients.\footnote{LSTMP baselines used 40d log-Mel filterbank analysis $+\Delta$ coeffs.} These inputs were normalised at the utterance level for mean and at the show-segment level for variance \cite{Woodland2015MGB}. All models were trained using the cross-entropy criterion and  frame-level shuffling used.

About 6k/9k decision tree clustered triphone tied-states along with GMM-HMM/DNN-HMM system training alignments were used for the 55h/275h training sets. The NewBob$^{+}$ learning rate scheduler \cite{Zhang2017Phd} was used to train all models with the setup for our previous MGB systems \cite{Woodland2015MGB}. An initial learning rate of $2\times 10^{-3}$ was used for all models and a 800 frame minibatch. \textit{L2}-regularisation was used was and tuned for the 55h systems but not tuned for the 275h systems. To give further context on the MGB3 data, the results are compared to a two layer projected LSTM (LSTMP) followed by an FC layer of same size before the output layer. The LSTMPs were implemented following \cite{Sak2014PLSTM}. For the 55h data set the width of the hidden layers is 500 and the projected vector size is 250. For the 275h dataset these were increased to 1000 and 500 respectively.

\section{Experimental Results}
\label{sec: ExpResults}

\subsection{Comparison of Kernels}
\label{ssec: Compkernel}
The different types of kernels given in Fig.~\ref{fig:Kernels} were investigated by using them for each kernel location in the TDNN shown in Fig.~\ref{fig:BaseTDNN}. The WERs for these structures are given in the first section of Table~\ref{table:55h}. The number of parameters in each of the models is kept roughly constant at 6.6M parameters, by adjusting the layer width. Therefore the layer widths for the TDNNs using the Standard-Kernel (TDNN) and ResNet-Kernel (ResNet-TDNN) are 653 and 500 respectively. The Double Kernel gives some WER reduction, but not as much as the ResNet Kernel. This might due to the lower minimum path through the network of the ResNet-TDNN. Given that the relative WER reduction (WERR) due to confusion network decoding is less for DT$^{\text{55h}}_1$ \& RT$^{\text{55h}}_1$ than for ST$^{\text{55h}}_1$ (1.9\% and 1.9\% compared to 2.7\%), it can be inferred that the deep kernels also sharpen the network output distributions.

\subsection{Appending Fully-Connected layers}
\label{ssec: ExpResults1}
To validate the improvement from adding deep kernels to the TDNN, the model is deepened in a simpler fashion. A number of FC layers are inserted between Kernel 4 and the output layer, which is equivalent to deepening Kernel 4.  FC layers were added to Kernel 4 until there is no further improvement in WER, again keeping the overall number of parameters in the network constant at 6.6M.\footnote{The effect of increasing the hidden layers' size is small as shown by  experiments using these structures trained with a hidden layers of 1000 nodes.}

The second section of Table \ref{table:55h} shows that TDNN architectures improve with increased depth of Kernel 4. The WER of a standard TDNN where Kernel 4 is four layers deep (TDNN+Deep) is similar to that of the ResNet-TDNN. However,  both changes  are complimentary and replacing Kernel 4 in the ResNet-TDNN by 3 FC layers (ResNet-TDNN+Deep) without any residual connection gives a further improvement. The WER reduction from the ResNet-Kernels increases with the 275h dataset as shown in Sec.~\ref{sseq: 275hExp}. The optimal number of additional FC layers will depend on the task and the data set. Hence, adding additional FC hidden layers to the TDNN using the CNN or the Grid-RNN was not investigated.

\begin{table}
\centerline{
\begin{tabular}{ll|ll}
\hline
ID & System & vit & cn\\\hline
ST$^{\text{55h}}_1$ & TDNN        & 33.6 & 32.7 \\
DT$^{\text{55h}}_1$ & Double-TDNN & 32.1 & 31.5 \\
RT$^{\text{55h}}_1$ & ResNet-TDNN & 31.1 & 30.5 \\\hline
ST$^{\text{55h}}_2$ & TDNN + 1 FC & 32.7 & 31.9 \\
ST$^{\text{55h}}_3$ & TDNN + 2 FC & 31.6 & 30.9 \\
ST$^{\text{55h}}_4$ & TDNN+Deep   & 31.2 & 30.5 \\
RT$^{\text{55h}}_2$ & ResNet-TDNN+Deep & 30.5 & 29.8 \\\hline
SC$^{\text{55h}}_1$ & CNN-TDNN & 31.6 & 31.0 \\
RC$^{\text{55h}}_1$ & CNN-ResNet-TDNN & 30.7 & 29.9 \\\hline
SG$^{\text{55h}}_1$ & Grid-RNN-TDNN & 31.2 & 30.5 \\
RG$^{\text{55h}}_1$ & Grid-RNN-ResNet-TDNN & 30.8 & 30.1\\
RG$^{\text{55h}}_2$ & FD-Grid-RNN-ResNet-TDNN & 30.2 & 29.6\\
RG$^{\text{55h}}_3$ & BD-FD-Grid-RNN-ResNet-TDNN & 29.7 & 29.0 \\\hline
L$^{\text{55h}}_1$ & 2L-LSTMP & 31.3 & 30.6 \\\hline
\end{tabular}
}
\caption{\it \%WERs for 55h systems on dev17b. Results are with a trigram LM and Viterbi decoding (vit) or CN decoding (cn).}
\label{table:55h}
\vspace{-5mm}
\end{table}

\subsection{Addition of Frequency Convolution}
The third section of Table \ref{table:55h} shows the improvements from adding frequency domain convolution to the TDNN structure with the Standard-Kernel (CNN-TDNN) and with the ResNet-Kernel (CNN-ResNet-TDNN). The frequency convolution used 100 filters per frequency band. The larger output size of the CNN layer results an increase in the total number of parameters. For the Standard-Kernel and the ResNet-Kernel, rel. WERRs of 5.2\% and 2.0\% (ST$^{\text{55h}}_1$ vs. SC$^{\text{55h}}_1$ and RT$^{\text{55h}}_1$ vs. RC$^{\text{55h}}_1$) respectively were achieved.

\subsection{Adding the Grid-RNN}
Section four of Table \ref{table:55h} shows the improvements from adding the Grid-RNN to the TDNN with the Standard-Kernel (Grid-RNN-TDNN) and with the ResNet-Kernel (Grid-RNN-ResNet-TDNN). The Grid-RNN had a width of 250 for the $\sigma$-RNN and of 500 for the \textit{Linear}-RNN. For the Standard-Kernel, a rel.~WERR of 6.7\% (ST$^{\text{55h}}_1$~vs.~SG$^{\text{55h}}_1$) was achieved. For the ResNet-Kernel the relative reduction is 1.3\% (RT$^{\text{55h}}_1$~vs.~RG$^{\text{55h}}_1$) which increases to 3.0\% as the parameters across the frequency domain are untied (RG$^{\text{55h}}_2$), and to 4.9\% for the bidirectional FD-Grid-RNN-ResNet-TDNN (RG$^{\text{55h}}_3$). This is an overall 11.3\% rel. improvement over the baseline TDNN (ST$^{\text{55h}}_1$~vs.~RG$^{\text{55h}}_3$). These experiments show the strength of the FD-Grid-RNN over the CNN. Besides the ability to model correlations between features in time and in frequency as discussed in previous work, the FD-Grid-RNN has many more parameters designated to the spectro-temporal modelling. In the CNN only 17.5K independent parameters are used for the convolution, a very small fraction of the total number of parameters, as opposed to the BD-FD-Grid-RNN with 1.4M parameters. At the same time the input to the first TDNN-Kernel is larger for the CNN. This shows an issue with the CNN, which is that it  only uses a very small number of parameters if  the input to the first TDNN-Kernel is kept at a be reasonable size.

\subsection{Further Experiments}
\label{sseq: 275hExp}
The performance of key modelling approaches was also tested using the larger 275h training set, and  the results shown in Table \ref{table: 275h}. Each of the models has a hidden layer width of 1000, except the $\sigma$-RNN in the Grid-RNNs, which is size 500. The deep kernels continue to give considerable improvement on the larger dataset. Further, the improvement due to the deep kernels scales better to the larger dataset than simply adding hidden layers: ResNet-TDNN+Deep has a 4\% relative lower WER than TDNN+Deep in comparison to 2\% on the smaller dataset (RT$^{\alpha}_2$~vs.~ST$^{\alpha}_4$).

\begin{table}[th]
\centerline{
\begin{tabular}{ll|ll}
  \hline
  ID & System &  vit & cn\\\hline
  ST$^{\text{275h}}_1$ & TDNN & 27.3 & 26.7 \\
  RT$^{\text{275h}}_1$ & ResNet-TDNN & 25.4 & 25.0 \\
  ST$^{\text{275h}}_4$ & TDNN+Deep   & 26.2 & 25.7 \\%
  RT$^{\text{275h}}_2$ & ResNet-TDNN+Deep &  25.1 & 24.7 \\ 
  RG$^{\text{275h}}_3$ & BD-FD-Grid-RNN-ResNet-TDNN & 24.6 & 24.3 \\
  L$^{\text{275h}}_1$ & 2L-LSTMP & 26.0 & 25.6 \\\hline
 \end{tabular}
}
\caption{\it \%WERs for 275h systems on dev17b. Results are with a trigram LM and Viterbi decoding (vit) or CN decoding (cn).}
\label{table: 275h}
\vspace{-5mm}
\end{table}

\section{Conclusion}
\label{sec:Conclusion}

In this paper, we presented different extensions to the \textit{sub-sampled} TDNN architecture. Deep Kernels for more abstract feature extraction as well as CNNs and 2D-RNN to reduce the spectro-temporal variation of the input feature. We propose a 2D-RNN architecture that does not rely on LSTMs and is complimentary to the TDNN architecture. We found that using both Deep Kernels and the 2D-RNN offers the results in the best performance. Overall, the combined structure yields a 9\% relative reduction in WER over the baseline TDNN architecture.

\bibliographystyle{IEEEbib}

\begin{thebibliography}{10}

\bibitem{waibel1989phoneme}
A.~Waibel, T.~Hanazawa, G.~Hinton, K.~Shikano, and K.~J Lang,
\newblock ``Phoneme recognition using time-delay neural networks,''
\newblock {\em IEEE Trans ASSP}, vol. 37, no. 3, pp. 328--339, 1989.

\bibitem{peddinti2015time}
V.~Peddinti, D.~Povey, and S.~Khudanpur,
\newblock ``A time delay neural network architecture for efficient modeling of
  long temporal contexts.,''
\newblock {\em Proc. Interspeech}, Dresden, 2015.

\bibitem{lecun1998gradient}
Y.~LeCun, L.~Bottou, Y.~Bengio, and P.~Haffner,
\newblock ``Gradient-based learning applied to document recognition,''
\newblock {\em Proceedings of the IEEE}, vol. 86, no. 11, pp. 2278--2324, 1998.

\bibitem{abdel2012applying}
O.~Abdel-Hamid, A.~Mohamed, H.~Jiang, and G.~Penn,
\newblock ``Applying convolutional neural networks concepts to hybrid
  {NN}-{HMM} model for speech recognition,''
\newblock {\em Proc. ICASSP}, Kyoto, 2012.

\bibitem{krizhevsky2012imagenet}
A.~Krizhevsky, I.~Sutskever, and G.E. Hinton,
\newblock ``Imagenet classification with deep convolutional neural networks,''
\newblock {\em Proc. NIPS}, Lake Tahoe, 2012.

\bibitem{Simonyan15}
K.~Simonyan and A.~Zisserman,
\newblock ``Very deep convolutional networks for large-scale image
  recognition,''
\newblock {\em Proc. ICLR}, San Diego, 2015.

\bibitem{he2016deep}
K.~He, X.~Zhang, S.~Ren, and J.~Sun,
\newblock ``Deep residual learning for image recognition,''
\newblock {\em Proc. CVPR}, Las Vegas, 2016.

\bibitem{lin2013network}
M.~Lin, Q.~Chen, and S.~Yan,
\newblock ``Network in network,''
\newblock {\em Proc. ICLR}, Scottsdale, 2013.

\bibitem{exploring-multidimensional-lstms-for-large-vocabulary-asr}
J.~Li, A.~Mohamed, G.~Zweig, and Y.~Gong,
\newblock ``Exploring multidimensional {LSTM}s for large vocabulary {ASR},''
\newblock {\em Proc. ICASSP}, Shanghai, 2016.

\bibitem{Tara2016ModelingTF}
T.~Sainath and B.~Li,
\newblock ``Modeling time-frequency patterns with {LSTM} vs. convolutional
  architectures for {LVCSR} tasks,''
\newblock {\em Proc. Interspeech}, San Francisco, 2016.

\bibitem{Li2017AcousticGoogleHome}
B.~Li, T.~Sainath, A.~Narayanan, J.~Caroselli, M.~Bacchiani, A.~Misra,
  I.~Shafran, H.~Sak, G.~Pundak, K.~Chin, K.C. Sim, R.J. Weiss, K.~Wilson,
  E.~Variani, C.~Kim, O.~Siohan, M.~Weintraub, E.~McDermott, R.~Rose, and
  M.~Shannon,
\newblock ``Acoustic modeling for {G}oogle {H}ome,''
\newblock {\em Proc. Interspeech}, Stockholm, 2017.

\bibitem{li2017reducing}
B.~Li and T.~Sainath,
\newblock ``Reducing the computational complexity of two-dimensional {LSTM}s,''
\newblock {\em Proc. Interspeech}, Stockholm, 2017.

\bibitem{waibel1989modular}
A.~Waibel,
\newblock ``Modular construction of time-delay neural networks for speech
  recognition,''
\newblock {\em Neural computation}, vol. 1, no. 1, pp. 39--46, 1989.

\bibitem{sainath2013deep}
T.N. Sainath, A.~Mohamed, B.~Kingsbury, and B.~Ramabhadran,
\newblock ``Deep convolutional neural networks for {LVCSR},''
\newblock {\em Proc. ICASSP}, Vancouver, 2013.

\bibitem{sainath2015convolutional}
T.N. Sainath, O.~Vinyals, A.~Senior, and H.~Sak,
\newblock ``Convolutional, long short-term memory, fully connected deep neural
  networks,''
\newblock {\em Proc. ICASSP}, Brisbane, 2015.

\bibitem{sercu2016very}
T.~Sercu, C.~Puhrsch, B.~Kingsbury, and Y.~LeCun,
\newblock ``Very deep multilingual convolutional neural networks for {LVCSR},''
\newblock {\em Proc. ICASSP}, Shanghai, 2016.

\bibitem{Qian2016VeryDeep}
Y.~Qian and P.C.~Woodland,
\newblock ``Very deep convolutional neural
networks for robust speech recognition,''
\newblock {\em Proc. SLT}, San Diego, 2016.

\bibitem{convolutional-neural-networks-for-speech-recognition-2}
O.~Abdel-Hamid, A.~Mohamed, H.~Jiang, L.~Deng, G.~Penn, and D.~Yu,
\newblock ``Convolutional neural networks for speech recognition,''
\newblock {\em IEEE/ACM Transactions on Audio, Speech, and Language
  Processing}, vol. 22, pp. 1533--1545, 2014.

\bibitem{hochreiter1997long}
S.~Hochreiter and J.~Schmidhuber,
\newblock ``Long short-term memory,''
\newblock {\em Neural computation}, vol. 9, no. 8, pp. 1735--1780, 1997.

\bibitem{Sak2014PLSTM}
H.~Sak, A.~Senior, and F.~Beaufays,
\newblock ``Long short-term memory recurrent neural network architectures for
  large scale acoustic modeling,''
\newblock {\em Proc. Interspeech}, Singapore, 2014.

\bibitem{peddinti2017low}
V.~Peddinti, Y.~Wang, D.~Povey, and S.~Khudanpur,
\newblock ``Low latency acoustic modeling using temporal convolution and
  {LSTM}s,''
\newblock {\em IEEE Signal Processing Letters}, 2017.

\bibitem{li2015lstm}
J.~Li, A.~Mohamed, G.~Zweig, and Y.~Gong,
\newblock ``{LSTM} time and frequency recurrence for automatic speech
  recognition,''
\newblock {\em Proc. ASRU}, Scottsdale, 2015.

\bibitem{kalchbrenner2015grid}
Nal Kalchbrenner, Ivo Danihelka, and Alex Graves,
\newblock ``Grid long short-term memory,''
\newblock {\em Proc. ICLR}, San Juan, 2016.

\bibitem{Saon2014Unfolded}
G.~Saon, H.~Soltau, A.~Emami, and M.~Picheny,
\newblock ``Unfolded recurrent neural networks for speech recognition,''
\newblock {\em Proc. Interspeech}, Singapore, 2014.

\bibitem{pascanu2013difficulty}
R.~Pascanu, T.~Mikolov, and Y.~Bengio,
\newblock ``On the difficulty of training recurrent neural networks,''
\newblock {\em Proc. ICML}, Atlanta, 2013.

\bibitem{Bell2015MGB}
P.~Bell, M.J.F. Gales, T.~Hain, J.~Kilgour, X.~Liu P.~Lanchantin, A.~McParland,
  S.~Renals, O.~Saz, M.~Wester, and P.C.~Woodland,
\newblock ``The {MGB} challenge: {E}valuating multi-genre broadcast media
  transcription,''
\newblock {\em Proc. ASRU}, Scottsdale, 2015.

\bibitem{MGB2017WEB}
``http://www.mgb-challenge.org,'' .

\bibitem{lanchatin2015g}
P.~Lanchantin, M.J.F. Gales, P.~Karanasou, X.~Liu, Y.~Qian, L.~Wang, P.C.
  Woodland, and C.~Zhang,
\newblock ``Selection of {M}ulti-{G}enre {B}roadcast data for the training of
  automatic speech recognition systems,''
\newblock {\em Proc. Interspeech}, San Francisco, 2016.

\bibitem{Richmond2010Combilex}
K.~Richmond, R.~Clark, and S.~Fitt,
\newblock ``On generating {C}ombilex pronunciations via morphological
  analysis,''
\newblock {\em Proc. Interspeech}, 2010.

\bibitem{Mangu2000CN}
L.~Mangu, E.~Brill, and A.~Stolcke,
\newblock ``Finding consensus in speech recognition: Word error minimization
  and other applications of confusion networks,''
\newblock {\em Computer Speech and Language}, vol. 14, pp. 373--400, 2000.

\bibitem{Evermann2000CN}
G.~Evermann and P.~Woodland,
\newblock ``Large vocabulary decoding and confidence estimation using word
  posterior probabilities,''
\newblock {\em Proc. ICASSP}, Istanbul, 2000.

\bibitem{Young2015HTK}
S.~Young, G.~Evermann, M.~Gales, T.~Hain, D.~Kershaw, X.~Liu, G.~Moore,
  J.~Odell, D.~Ollason, D.~Povey, A.~Ragni, V.~Valtchev, P.~Woodland, and
  C.~Zhang,
\newblock ``The {HTK} book (for {HTK} version 3.5),''
\newblock {\em Cambridge University Engineering Department}, 2015.

\bibitem{zhang2015general}
C.~Zhang and P.C. Woodland,
\newblock ``A general artificial neural network extension for {HTK},''
\newblock {\em Proc. Interspeech}, Dresden, 2015.

\bibitem{Woodland2015MGB}
P.C. Woodland, X.~Liu, Y.~Qian, C.~Zhang, P.~Karanasou M.J.F.~Gales,
  P.~Lanchantin, and L.~Wang,
\newblock ``Cambridge university transcription systems for the {M}ulti-{G}enre
  {B}roadcast challenge,''
\newblock {\em Proc. ASRU}, Scottsdale, 2015.

\bibitem{Zhang2017Phd}
C.~Zhang,
\newblock {\em Joint Training Methods for Tandem and Hybrid Speech Recognition
  Systems using Deep Neural Networks},
\newblock Ph.D. thesis, University of Cambridge, Cambridge,UK, 2017.

\end{thebibliography}

\end{document}